\pdfoutput=1

\documentclass[11pt]{article}

\usepackage{emnlp2021}

\usepackage{microtype}
\usepackage{graphicx}
\usepackage{subfigure}
\usepackage{booktabs} 
\usepackage{amsmath}
\usepackage{amssymb}
\usepackage{amsfonts}
\usepackage{multirow}
\usepackage{verbatim}
\usepackage{caption}
\usepackage{enumitem}
\usepackage{tablefootnote}

\usepackage{hyperref}
\usepackage{cleveref}

\DeclareMathOperator{\softmax}{softmax}

\DeclareMathOperator{\layernorm}{LayerNorm}

\usepackage[utf8]{inputenc}

\usepackage{microtype}

\title{Do Transformer Modifications Transfer Across Implementations and Applications?}

\author{
Sharan Narang\thanks{ \, Correspondence to \texttt{sharannarang@google.com}}\\
\And
Hyung Won Chung\\
\And
Yi Tay\\
\And
William Fedus\\
\AND
Thibault Fevry\thanks{ \, Work completed while at Google}\\
\And
Michael Matena\footnotemark[2]\\
\And
Karishma Malkan\footnotemark[2]\\
\And
Noah Fiedel\\
\AND
Noam Shazeer\\
\And
Zhenzhong Lan\footnotemark[2]\\
\And
Yanqi Zhou\\
\And
Wei Li\\
\AND
Nan Ding \\
\And
Jake Marcus\\
\And
Adam Roberts\\
\And
Colin Raffel\footnotemark[2]
}

\begin{document}
\maketitle
\begin{abstract}

The research community has proposed copious modifications to the Transformer architecture since it was introduced over three years ago, relatively few of which have seen widespread adoption.
In this paper, we comprehensively evaluate many of these modifications in a shared experimental setting that covers most of the common uses of the Transformer in natural language processing.
Surprisingly, we find that most modifications do not meaningfully improve performance.
Furthermore, most of the Transformer variants we found beneficial were either developed in the same codebase that we used or are relatively minor changes.
We conjecture that performance improvements may strongly depend on implementation details and correspondingly make some recommendations for improving the generality of experimental results.
\end{abstract}

\section{Introduction}

Much of the empirical success of deep learning can be attributed to advances in methods for building and training neural networks.
These advances include improved optimizers \cite{sutskever2013importance,hinton2012lecture,kingma2014adam,shazeer2018adafactor}, regularization schemes \cite{srivastava2014dropout,zhang2017mixup,neelakantan2015adding}, and model architectures \cite{he2016deep,hochreiter1997long,vaswani2017attention}.
An aspiration underlying much of this work is that an improvement to a particular machine learning pipeline will yield equal-or-better performance on any task that the pipeline is applicable to.
For example, residual connections in convolutional networks \cite{he2016deep} are designed to ideally improve performance on any task where these models are applicable (image classification, semantic segmentation, etc.).
In practice, when proposing a new improvement, it is impossible to test it on every applicable downstream task, so researchers must select a few representative tasks to evaluate it on.
However, the proposals that are ultimately adopted by the research community and practitioners tend to be those that reliably improve performance across a wide variety of tasks ``in the wild".

The Transformer architecture \cite{vaswani2017attention} is an example of a seminal improvement in the field of deep learning.
Currently, the Transformer is the \textit{de facto} architecture of choice for processing sequential data and is starting to be applied to vision applications (e.g.~\citet{dosovitskiy2020image}).
Since being introduced three years ago, many modifications to the Transformer architecture have been proposed.
However, the most widely-used applications of the Transformer architecture (e.g.~\citet{devlin2018bert,yang2019xlnet,radford2018improving,raffel2019exploring}) incorporate few of these modifications.
Instead, the standard practice is to use a slightly-modified version of the originally-proposed Transformer.
One possible explanation for this is that the originally-proposed Transformer architecture was near-perfect, and there wasn't much that could be done to improve it.
This is in contrast to, for example, convolutional neural networks, which have continually evolved over the past few decades (e.g.\ the replacement of pooling with striding \cite{springenberg2014striving}, fully-connected layers with convolutional layers \cite{lin2013network}, the addition of normalization \cite{ioffe2015batch} and residual connections \cite{he2016deep}, etc.).
Another possible explanation is that the modifications proposed to the Transformer do not ``generalize'' across applications, i.e.\ the modifications only help on the limited experimental setting considered when the modification was proposed, and/or rely on specific details that are not common across implementations of the Transformer.

The main goal of this paper is to try to determine why most modifications proposed to the Transformer have not seen widespread adoption.
To answer this question, we reimplemented and evaluated a wide variety of Transformer variants on a suite of tasks that Transformers are commonly applied to.
Our main finding is that many Transformer modifications do not result in improved performance in our experimental setting.
Moreover, those variants that did yield better performance tended to be those that were quite small changes and/or were developed in the codebase where we carried out our evaluation.
This suggests to us the possibility that Transformer modifications exhibit a surprising lack of generalization across different implementations and tasks.


\vspace{-5pt}
\section{Modifications}
\label{sec:modifications}

\vspace{-5pt}

In this section, we enumerate all of the architectural modifications we consider. For a description of the Transformer architecture, refer to the \cref{sec:background}.

Due to space constraints, we are seldom able to thoroughly define each specific modification. Moreover, we limit our study to the encoder-decoder architecture.
Please refer to the original sources for each modification for additional details.


\vspace{-5pt}
\subsection{Activations}
\label{sec:activations}
We consider various activation functions to replace the ReLU in the feedforward network block. The activation functions that we explored are: (1) GeLU \citep{hendrycks2016gaussian}, (2) Swish \citep{ramachandran2017searching}, (3) Exponential Linear Units (ELU) \citep{clevert2015fast}, (4) Scaled exponential linear units (SeLU) \citep{klambauer2017self},  (5) Sigmoid and (6) Softplus.
 We also explore ``Gated Linear Unit'' (GLU) variants \citep{dauphin2017language,shazeer2020glu} which compose two linear transformations together in an element-wise fashion, i.e. $F_1(x) \odot \sigma(F_2(x))$ where $\sigma$ is an activation function and $F_1$ and $F_2$ are separate learned affine transformations. We explore modifying $\sigma$ to be sigmoid activations (denoted as GLU), ReLU activations (denoted as ReGLU), GeLU activations (denoted as GeGLU) or to be a standard linear transformation (no activation, denoted as LiGLU).









\vspace{-5pt}
\subsection{Normalization}
\label{sec:normalization}
We explored ``RMS (root-mean-square) norm'' \citep{zhang2019root} as an alternative to layer normalization as well as the Rezero \citep{bachlechner2020rezero} initialization scheme, including combining Rezero with Layer Norm and RMS Norm. We also explored the Fixup \citep{zhang2019fixup} initialization scheme which tries to solve the vanishing/exploding gradient problem by rescaling the initializations.

\vspace{-5pt}
\subsection{Depth}
\label{sec:depth}
We explored the trade-offs between the width of the feedforward subblocks ($d_{\rm{ff}}$) and depth ($L$). In order to ensure fair comparison, we scale $d_{\rm{ff}}$ and the number of heads ($H$) in order to keep the total number of parameters constant when changing the depth.

\subsection{Embeddings}
\label{sec:embeddings}

The Transformer model includes multiple weight matrices of shape of $d_{\rm{model}} \times d_{\rm{vocab}}$: one at the input of the encoder, one at the input of the decoder, and one at the output of the decoder. \citet{chung2021rethinking} showed the benefits of untying the embeddings for the encoder-only models. We extend the analysis and explore various ways of sharing these parameters: tying only encoder input and decoder input embeddings, tying only decoder input and output embeddings, and untying all the embeddings. 

In addition, we explored factorizing the embedding matrix into two smaller matrices \cite{lan2019albert}. In other words, the embedding matrix of size $[d_{\rm{model}}, d_{\rm{vocab}}]$ is factored into $[d_{\rm{model}}, d_{\rm{inner}}]$ and $[d_{\rm{inner}}, d_{\rm{vocab}}]$. We tried both untied and tied decoder embeddings while encoder and decoder embeddings are shared.

The last technique we explored for the embeddings is the ``Adaptive input embeddings” by~\citet{baevski2018adaptive}. Vocabulary items are clustered based on their frequencies. A cluster with more frequent ones has a larger embedding dimension. The embedding vectors are projected to the same dimension and concatenated.

\subsection{Parameter sharing}
\label{sec:param-sharing}
We also explored sharing the parameters of the Transformer layers inspired by the ``ALBERT'' model of \citet{Lan2020ALBERT}. Each subblock (e.g., self-attention) has a unique set of weights shared across all $l$ layers. Following~\citet{Lan2020ALBERT}, we factorized the embeddings (denoted as ``Factorized embeddings") in addition to the parameter sharing. Note that these models have untied softmax and vocabulary embeddings in the decoder; we also tried tying them (denoted as ``Shared embeddings").
Finally, we experimented with applying the parameter sharing to the encoder and decoder separately.

\subsection{Softmax}
\label{sec:softmax}
Our work considers variations to the softmax computation that produces the final probability distribution as computed by the last layer embedding.
Adaptive softmax ~\citep{joulin2017efficient} uses the natural imbalance in word distributions \citep{zipf1949human} to form clusters in a hierarchical model, which minimizes computation time.
In the original implementation, each cluster is permitted to have a different capacity and the size of the representations for rare words is reduced via a projection matrix.
We consider the original variant, as well as a version that ablates the projection operation.
Mixture of Softmaxes (MoS) ~\cite{yang2017breaking} improves the expressiveness of a single softmax operation by instead computing a linear combination over softmaxes, each weighted by learned coefficients.

\subsection{Architectures}
\label{sec:architectures}

\textbf{Transparent Attention} \hspace{0.2cm}  One type of attention variant we experiment with is Transparent Attention \citep{bapna2018training}. Transparent attention~\citep{bapna2018training} creates weighted residual connections along encoder depth to facilitate gradient flow. In \cref{sec:pos-embedding}, we experiment with additional attention variants.

\textbf{Evolved Transformer} \hspace{0.2cm} The Evolved Transformer \citep{so2019evolved} was designed via evolution-based architecture search \citep{real2019regularized} where the initial population was seeded with the original Transformer.
The search space generalizes the one followed in NASNet \citep{zoph2018learning}, but extended to be able to represent the Transformer.

\textbf{Synthesizer variants} \hspace{0.2cm} We explore the factorized, dense, and random Synthesizer variants from \citet{tay2020synthesizer}, where self-attention is replaced with ``synthetic attention'' patterns. We denote ``plus'' when dot product attention is additively combined with the synthetic attention and \textit{plus alpha} to denote when a scalar $\alpha$ is used to interpolate between synthetic and dot product attention.

\textbf{Funnel Transformer} \hspace{0.2cm} Funnel Transformer progressively reduces the sequence length in order to efficiently encode the input sequence~\citep{Dai2020Funnel}. We only applied this reduction to the encoder.

\textbf{Lightweight and Dynamic convolutions} \hspace{0.2cm} Lightweight convolution~\citep{wu2018pay} is a special case of a depth-wise convolution. It shares the weights of every subsequent number of $m$ channels where $m$ is a hyperparameter and normalizes the weights across the filter dimension. For a Transformer model, the depth dimension corresponds to $d_{\rm{model}}$.
Dynamic convolution~\citep{wu2018pay} uses kernels that are functions of the input at the current time step. Following \citet{wu2018pay}, we compute the kernels as a simple linear function of the layer input.

\textbf{Sparse Expert Transformers} \hspace{0.2cm} Mixture of Experts (MoE) Transformer~\citep{shazeer2018mesh,lepikhin2020gshard} and Switch Transformer~\citep{fedus2021switch} both replace the feedforward network with sparsely activated experts layers.
The result is an example of adaptive computation where parameters (expert FFNs) are selected for each specific token.
This provides a way of scaling up the parameter count of a model independently from the FLOPs required for a forward pass.
Some variants in \citet{fedus2021switch} consider sparse self-attention layers as well, but we only consider the primary variant here.

\textbf{Product Key Memory} \hspace{0.2cm} Similar to the expert model designs, product key memory networks~\citep{lample2019large} process inputs adaptively, selecting sparse values.
In contrast, the mechanism of sparse computation isn't done via learned routing, but instead by an efficient $k$-nearest neighbor weighted sum. 

\textbf{Universal Transformer} \hspace{0.2cm} Similar to block sharing, the Universal Transformer \cite{dehghani2018universal} applies the same Transformer ``block'' over and over again to the input sequence.
However, instead of applying it a fixed number of times, it recurrently refines the representation for each token until a halting mechanism (based on Adaptive Computation Time \cite{graves2016adaptive}) is triggered.

\section{Experiments}
\label{sec:experiments}


In order to study the impact of each of the modifications described in \cref{sec:modifications}, we conduct a systematic study by comparing a baseline model to each modification while holding the task, hyperparameters, optimizer, and either the parameter count or FLOP budget (floating point operations per second) constant. We use the original Transformer model as our baseline model with two modifications: First, we apply layer normalization before the self-attention and feedforward blocks instead of after.
This small change has been unanimously adopted by all current Transformer implementations because it leads to more effective training \citep{baevski2018adaptive,xiong2020layer}.
Secondly, we use relative attention with shared biases (as used in \citet{raffel2019exploring}) instead of sinusoidal positional embeddings, which makes it easier to train the model.
Our baseline model is a standard encoder-decoder with 12 layers in the encoder and decoder. The feedforward network in each layer consists of a dense layer with dimension of $d_{\mathrm{ff}} = 3072$. All attention mechanisms have 12 heads and ``key" and ``value" matrices have a dimension of  $d_{\mathrm{kv}} = 64$. All other sublayers have a dimension of $d_{\mathrm{model}} = 768$ resulting in 223 million parameters in the model. We refer to this model as the ``Vanilla Transformer". 

We consider two experimental settings for evaluating the performance of each modification: Transfer learning based on the T5 \cite{raffel2019exploring} and supervised machine translation on the WMT'14 English-German translation.

For transfer learning, we copy the methodology used by the T5 model, proposed in \citet{raffel2019exploring}.
For full details of this experimental setup, please refer to \citet{raffel2019exploring}.
We pre-train encoder-decoder models in a self-supervised manner using the ``span corruption'' masked language modeling objective \cite{taylor1953cloze,fedus2018maskgan,devlin2018bert} on the C4 dataset.
We run experiments on version 2.3.1 of the C4 dataset available in TensorFlow Datasets\footnote{\url{https://www.tensorflow.org/datasets/catalog/c4}}.
We pre-train each architecture variant for $524,288$ steps with batches of $65,536$ tokens.
As in T5, we use Adafactor \citep{DBLP:journals/corr/abs-1804-04235} for optimization and an inverse square root learning rate schedule during pre-training.
We use a maximum sequence length of 512 for both the inputs and targets during pre-training. 
To evaluate the performance of pre-trained models, we compute the perplexity on a held-out portion of the C4 dataset for each pre-trained model, with the expectation that improvements in perplexity will correlate with performance on fine-tuned tasks.
To capture the inter-run variance on these models, we run each model $5$ times for $65,536$ steps ($\frac{1}{8}$th of the total pre-training steps).
We report the mean and standard deviation of the loss (log perplexity) on held-out data of these five experiments and also report the final loss at the end of pre-training ($524,288$ steps). We do not use any regularization during pre-training.

In the transfer learning setting, after pre-training we fine-tune each model on three different tasks: the SuperGLUE \citep{superglue} natural language understanding meta-benchmark, the XSum \citep{xsum-2018} abstractive summarization dataset, and the closed-book variant \cite{roberts2020much} of the WebQuestions \citep{berant-etal-2013-semantic} question-answering task.
With these tasks, we hope to capture a broad variety of NLP problems including language understanding and classification, language generation, and knowledge internalization.
For SuperGLUE and XSum, each model is fine-tuned for $262{,}144$ steps.
Since the WebQuestions dataset is much smaller, we fine-tune the model for only $30{,}000$ steps.
We use a constant learning rate of $0.0005$ with a linear warm-up of $20,000$ steps.
Similar to pre-training, each batch contains $65,536$ tokens.
We save a checkpoint every $2,500$ steps ($1,000$ steps for WebQuestions) and report results on the model checkpoint corresponding to the highest validation performance. We use a dropout of $0.1$ during fine-tuning for all the tasks.
All results are reported on the validation split of each dataset.
For SuperGLUE, we report the average score across all tasks in the benchmark.
We report ROUGE-2 \citep{lin-2004-rouge} for XSum and accuracy for WebQuestions. 

For supervised training on the WMT'14 English to German translation task~\cite{bojar-EtAl:2014:W14-33}, we use the same model and batch size as for the transfer learning setting.
We train for a total of $150{,}000$ steps.
We use the same data splits as were used in \cite{vaswani2017attention} and report the BLEU score of the highest-scoring checkpoint on the validation set. We use a vocabulary of 37,000 tokens learned by Byte Pair Encoding~\citep{sennrich-etal-2016-neural} for supervised training as opposed to 32,000 tokens (created using SentencePiece~\cite{kudo2018sentencepiece}) for the transfer learning experiments.

To compare the efficiency of the model, we also report the total number of parameters, the total number of floating point operations, and the measured steps per second in the pre-training experiments.
Reporting these parameters can help us understand the trade-off between quality and efficiency.
For each architectural modification, we attempt to keep either the parameter count or total operations in the model approximately the same to perform a fair comparison with the baseline model. 

All hyperparameters are held constant for each architectural variant across pre-training and fine-tuning. However, we found that certain architectural (Rezero and Fixup) variants achieved significantly lower negative log perplexity than the baseline model with the Adafactor optimizer. Therefore, we use the Adam optimizer \citep{kingma2014adam} for these variants. For pre-training, we use an inverse square root learning rate schedule with a linear warm-up of $4,000$ steps. For fine-tuning, we use a constant learning rate of $5e-5$ with a linear warm-up of $20,000$ steps. We provide details of certain modifications in \cref{sec:implementation-details}.

All experiments are run using the T5 library \footnote{\url{https://github.com/google-research/text-to-text-transfer-transformer}} on ``slices" of Cloud TPU Pods. All model variants are implemented in the Mesh TensorFlow library \citep{shazeer2018mesh}. 

\subsection{Results}
\label{sec:results}

The results for all model variants are shown in \cref{tab:rel-attn-results}. The vanilla Transformer achieves a SuperGLUE average of $70.97$ and a BLEU score of $26.62$ on WMT14 EnDe. This is comparable with the scores achieved by the equivalently-sized T5-Base model \citet{raffel2019exploring} and similarly-sized Transformer-Big from \citet{vaswani2017attention}, which confirms that our baseline is reasonable. As mentioned earlier, each variant has approximately the same number of parameters or total operations as the vanilla Transformer, with the following exceptions: For the Universal Transformer, the total number of operations is approximately 4$\times$ the baseline model. Since the Universal Transformer model is already significantly smaller than the baseline model, it would not be fair to shrink the model even further to match the number of operations with the baseline. Product key memories \cite{lample2019large} should only slightly increase FLOPs over the vanilla Transformer, but the total number of operations is artificially extremely high due to an inefficient implementation in Mesh Tensorflow.

We find that several activation functions improve performance over the ReLU activation. Specifically, SwiGLU and GeGLU improve performance on pre-training, fine-tuning, and supervised training without sacrificing any efficiency in terms of speed. Replacing layer normalization with RMS normalization yields improvements while also improving training speed. Our experiments with varying the depth of the model indicate that deeper models tend to outperform shallower ones with a fixed parameter count. However, these deeper models are also more compute-intensive and therefore slower than their shallower counterparts. Sharing of parameters across layers tends to hurt performance. Interestingly, untying the encoder/decoder embeddings improve performance with only a modest increase in parameter count. Using mixture of softmaxes does improve performance but is almost $40\%$ slower than the vanilla Transformer. 

Among the different architectures, we find that two of the synthesizer variants are beneficial. Switch Transformer, mixture of experts, and product key memories all improve performance with significantly more parameters than the baseline model. However, these implementations only use a subset of the parameters during each step, so they are roughly equivalent to the vanilla Transformer in total number of operations. Surprisingly, all the other architecture variants generally performed poorly.

Overall, we found that most of the beneficial modifications conferred improvements across pre-training, fine-tuning, and supervised training, though a few variants (e.g.\ transparent attention, Synthesizer-random, fixup) harmed performance for transfer learning but not for WMT'14 EnDe. The modifications that led to significant improvements tended to fall into one of three buckets: relatively minor changes (i.e., activation functions, normalization and untying embedding matrices); those that increase parameter count (i.e., Switch Transformer, product key memory) or are slower~(i.e., mixture of softmaxes, deeper models); or those that were originally invented in the Mesh TensorFlow codebase that we use for our experiments (i.e., mixture of experts, switch Transformer, synthesizer). To further ensure the correctness of the various architecture modifications, we reached out to authors of 12 techniques to review our implementation and provide their feedback and received responses from 6 of them. All of the authors who responded confirmed that our re-implementation was correct.


\begin{table*}[ht!]
    \centering
    \resizebox{\textwidth}{!}{%
    \begin{tabular}{p{4.5cm}cccccccc|c}
    \toprule
    \textbf{Model} & \textbf{Params} & \textbf{Ops} & \textbf{Step/s} & \textbf{Early loss} & \textbf{Final loss} & \textbf{SGLUE} & \textbf{XSum} & \textbf{WebQ} & \textbf{WMT EnDe} \\
    \midrule
    Vanilla Transformer & $223M$ & $11.1T$ & $3.50$ & $2.182 \pm 0.005$ & $1.838$ & $71.66$ & $17.78$ & $23.02$ & $26.62$ \\ 
    \midrule
    GeLU &     $223M$ & $11.1T$ & $3.58$ & $2.179 \pm 0.003$ & $1.838$ & $\textbf{75.79}$ & $\textbf{17.86}$ & $\textbf{25.13}$ & $26.47$ \\
    Swish &    $223M$ & $11.1T$ & $3.62$ & $2.186 \pm 0.003$ & $1.847$ & $\textbf{73.77}$ & $17.74$ & $\textbf{24.34}$ & $\textbf{26.75}$\\  
    ELU &      $223M$ & $11.1T$ & $3.56$ & $2.270 \pm 0.007$ & $1.932$ & $67.83$ & $16.73$ & $23.02$ & $26.08$\\    
    GLU &      $223M$ & $11.1T$ & $3.59$ & $2.174 \pm 0.003$ & $\textbf{1.814}$ & $\textbf{74.20}$ & $\textbf{17.42}$ & $24.34$ & \textbf{27.12} \\ 
    GeGLU &    $223M$ & $11.1T$ & $3.55$ & $2.130 \pm 0.006$ & $\textbf{1.792}$ & $\textbf{75.96}$ & $\textbf{18.27}$ & $\textbf{24.87}$ & $\textbf{26.87}$ \\ 
    ReGLU &    $223M$ & $11.1T$ & $3.57$ & $2.145 \pm 0.004$ & $\textbf{1.803}$ & $\textbf{76.17}$ & $\textbf{18.36}$ & $\textbf{24.87}$ & $\textbf{27.02}$ \\ 
    SeLU &     $223M$ & $11.1T$ & $3.55$ & $2.315 \pm 0.004$ & $1.948$ & $68.76$ & $16.76$ & $22.75$ & $25.99$ \\  
    SwiGLU &   $223M$ & $11.1T$ & $3.53$ & $2.127 \pm 0.003$ & $\textbf{1.789}$ & $\textbf{76.00}$ & $\textbf{18.20}$ & $\textbf{24.34}$ & $\textbf{27.02}$ \\ 
    LiGLU &    $223M$ & $11.1T$ & $3.59$ & $2.149 \pm 0.005$ & $\textbf{1.798}$ & $\textbf{75.34}$ & $\textbf{17.97}$ & $\textbf{24.34}$ & $26.53$ \\ 
    Sigmoid &  $223M$ & $11.1T$ & $3.63$ & $2.291 \pm 0.019$ & $1.867$ & $\textbf{74.31}$ & $17.51$ & $23.02$ & $26.30$ \\  
    Softplus & $223M$ & $11.1T$ & $3.47$ & $2.207 \pm 0.011$ & $1.850$ & $\textbf{72.45}$ & $17.65$ & $\textbf{24.34}$ & \textbf{26.89}\\  
    \midrule
    RMS Norm           & $223M$ & $11.1T$ & $3.68$ & $2.167 \pm 0.008$ & $\textbf{1.821}$ & $\textbf{75.45}$ & $\textbf{17.94}$ & $\textbf{24.07}$ & \textbf{27.14} \\
    Rezero             & $223M$ & $11.1T$ & $3.51$ & $2.262 \pm 0.003$ & $1.939$ & $61.69$ & $15.64$ & $20.90$ & $26.37$ \\
    Rezero + LayerNorm & $223M$ & $11.1T$ & $3.26$ & $2.223 \pm 0.006$ & $1.858$ & $70.42$ & $17.58$ & $23.02$ & $26.29$ \\
    Rezero + RMS Norm  & $223M$ & $11.1T$ & $3.34$ & $2.221 \pm 0.009$ & $1.875$ & $70.33$ & $17.32$ & $23.02$ & $26.19$ \\
    Fixup              & $223M$ & $11.1T$ & $2.95$ & $2.382 \pm 0.012$ & $2.067$ & $58.56$ & $14.42$ & $23.02$ & $26.31$\\ 
    \midrule
    24 layers, $d_{\rm{ff}} = 1536, H = 6$ & $224M$ & $11.1T$ & $3.33$ & $2.200 \pm 0.007$ & $1.843$ & $\textbf{74.89}$ & $17.75$ & $\textbf{25.13}$ & $\textbf{26.89}$\\
    18 layers, $d_{\rm{ff}} = 2048, H = 8$ & $223M$ & $11.1T$ & $3.38$ & $2.185 \pm 0.005$ & $\textbf{1.831}$ & $\textbf{76.45}$ & $16.83$ & $\textbf{24.34}$ & $\textbf{27.10}$\\
    8 layers, $d_{\rm{ff}} = 4608, H = 18$  & $223M$ & $11.1T$ & $3.69$ & $2.190 \pm 0.005$ & $1.847$ & $\textbf{74.58}$ & $17.69$ & $\textbf{23.28}$ & $\textbf{26.85}$\\
    6 layers, $d_{\rm{ff}} = 6144, H = 24$  & $223M$ & $11.1T$ & $3.70$ & $2.201 \pm 0.010$ & $1.857$ & $\textbf{73.55}$ & $17.59$ & $\textbf{24.60}$ & $\textbf{26.66}$\\ 
    \midrule
    Block sharing                     & $65M$ & $11.1T$ & $3.91$ & $2.497 \pm 0.037$ & $2.164$ & $64.50$ & $14.53$ & $21.96$  & $25.48$ \\
    \hspace*{0.1cm} + Factorized embeddings           & $45M$ & $9.4T$ & $4.21$ & $2.631 \pm 0.305$ & $2.183$ & $60.84$ & $14.00$ & $19.84$ & $25.27$ \\
    \hspace*{0.1cm} + Factorized \& shared embeddings & $20M$ & $9.1T$ & $4.37$ & $2.907 \pm 0.313$ & $2.385$ & $53.95$ & $11.37$ & $19.84$ & $25.19$ \\
    Encoder only block sharing        & $170M$ & $11.1T$ & $3.68$ & $2.298 \pm 0.023$ & $1.929$ & $69.60$ & $16.23$ & $23.02$ & $26.23$ \\
    Decoder only block sharing        & $144M$ & $11.1T$ & $3.70$ & $2.352 \pm 0.029$ & $2.082$  & $67.93$ & $16.13$ & $\textbf{23.81}$ & $26.08$ \\ 
    \midrule
    Factorized Embedding                 & $227M$ & $9.4T$  & $3.80$ & $2.208 \pm 0.006$ & $1.855$ & $70.41$ & $15.92$ & $22.75$ & $26.50$ \\
    Factorized \& shared embeddings      & $202M$ & $9.1T$  & $3.92$ & $2.320 \pm 0.010$ & $1.952$ & $68.69$ & $16.33$ & $22.22$ & $26.44$ \\
    Tied encoder/decoder input embeddings & $248M$ & $11.1T$ & $3.55$ & $2.192 \pm 0.002$ & $1.840$ & $\textbf{71.70}$ & $17.72$ & $\textbf{24.34}$ & $26.49$ \\
    Tied decoder input and output embeddings  & $248M$ & $11.1T$ & $3.57$ & $2.187 \pm 0.007$ & $\textbf{1.827}$ & $\textbf{74.86}$ & $17.74$ & $\textbf{24.87}$ & $\textbf{26.67}$\\
    Untied embeddings & $273M$ & $11.1T$ & $3.53$ & $2.195 \pm 0.005$ & $\textbf{1.834}$ & $\textbf{72.99}$ & $17.58$ & $\textbf{23.28}$  & $26.48$ \\    
    Adaptive input embeddings                      & $204M$ & $9.2T$  & $3.55$ & $2.250 \pm 0.002$ & $1.899$ & $66.57$ & $16.21$ & $\textbf{24.07}$ & $\textbf{26.66}$ \\
    \midrule
    Adaptive softmax                    & $204M$ & $9.2T$  & $3.60$ & $2.364 \pm 0.005$ & $1.982$ & $\textbf{72.91}$ & $16.67$ & $21.16$ & $25.56$ \\
    Adaptive softmax without projection & $223M$ & $10.8T$ & $3.43$ & $2.229 \pm 0.009$ & $1.914$ & $\textbf{71.82}$ & $17.10$ & $23.02$ & $25.72$ \\
    Mixture of softmaxes                  & $232M$ & $16.3T$ & $2.24$ & $2.227 \pm 0.017$ & $\textbf{1.821}$ & $\textbf{76.77}$ & $17.62$ & $22.75$ & $\textbf{26.82}$ \\ 
    \midrule    
    Transparent attention           & $223M$ & $11.1T$ & $3.33$ & $2.181 \pm 0.014$ & $1.874$ & $54.31$ & $10.40$ & $21.16$ & $\textbf{26.80}$ \\
    Dynamic convolution             & $257M$ & $11.8T$ & $2.65$ & $2.403 \pm 0.009$ & $2.047$ & $58.30$ & $12.67$ & $21.16$ & $17.03$ \\  
    Lightweight convolution         & $224M$ & $10.4T$ & $4.07$ & $2.370 \pm 0.010$ & $1.989$ & $63.07$ & $14.86$ & $23.02$ & $24.73$\\    
    Evolved Transformer             & $217M$ &  $9.9T$ & $3.09$ & $2.220 \pm 0.003$ & $1.863$ & $\textbf{73.67}$ & $10.76$ & $\textbf{24.07}$ & $26.58$\\ 
    Synthesizer (dense)       & $224M$ & $11.4T$ & $3.47$ & $2.334 \pm 0.021$ & $1.962$ & $61.03$ & $14.27$ & $16.14$ & $\textbf{26.63}$\\ 
    Synthesizer (dense plus)        & $243M$ & $12.6T$ & $3.22$ & $2.191 \pm 0.010$ & $1.840$ & $\textbf{73.98}$ & $16.96$ & $\textbf{23.81}$ & $\textbf{26.71}$\\ 
    Synthesizer (dense plus alpha)  & $243M$ & $12.6T$ & $3.01$ & $2.180 \pm 0.007$ & $\textbf{1.828}$ & $\textbf{74.25}$ & $17.02$ & $\textbf{23.28}$ & $26.61$ \\  
    Synthesizer (factorized)        & $207M$ & $10.1T$ & $3.94$ & $2.341 \pm 0.017$ & $1.968$  & $62.78$ & $15.39$ & $\textbf{23.55}$ &  $26.42$\\ 
    Synthesizer (random)            & $254M$ & $10.1T$ & $4.08$ & $2.326 \pm 0.012$ & $2.009$ & $54.27$ & $10.35$ & $19.56$  &  $26.44$ \\ 
    Synthesizer (random plus)       & $292M$ & $12.0T$ & $3.63$ & $2.189 \pm 0.004$ & $1.842$ & $\textbf{73.32}$ & $17.04$ & $\textbf{24.87}$ &  $26.43$\\  
    Synthesizer (random plus alpha) & $292M$ & $12.0T$ & $3.42$ & $2.186 \pm 0.007$ & $\textbf{1.828}$ & $\textbf{75.24}$ & $17.08$ & $\textbf{24.08}$ & $26.39$ \\
    Universal Transformer           & $84M$  & $40.0T$ & $0.88$ & $2.406 \pm 0.036$ & $2.053$ & $70.13$ & $14.09$ & $19.05$ & $23.91$ \\ 
    Mixture of experts              & $648M$ & $11.7T$ & $3.20$ & $2.148 \pm 0.006$ & $\textbf{1.785}$ & $\textbf{74.55}$ & $\textbf{18.13}$ & $\textbf{24.08}$  & $\textbf{26.94}$ \\ 
    Switch Transformer             & $1100M$ & $11.7T$ & $3.18$ & $2.135 \pm 0.007$ & $\textbf{1.758}$ & $\textbf{75.38}$ & $\textbf{18.02}$ & $\textbf{26.19}$ & $\textbf{26.81}$ \\  
    Funnel Transformer              & $223M$ & $1.9T$ & $4.30$  & $2.288 \pm 0.008$ & $1.918$ & $67.34$ & $16.26$ & $22.75$  & $23.20$\\ 
    Weighted Transformer            & $280M$ & $71.0T$ & $0.59$ & $2.378 \pm 0.021$ & $1.989$ & $69.04$ & $16.98$ & $23.02$ & $26.30$ \\ 
    Product key memory             & $421M$ & $386.6T$ & $0.25$ & $2.155 \pm 0.003$ & $\textbf{1.798}$ & $\textbf{75.16}$ & $17.04$ & $\textbf{23.55}$ & $\textbf{26.73}$ \\  
    \bottomrule
    \end{tabular}}
    \caption{Results for all architecture variants. The baseline model is the vanilla Transformer with relative attention. The early loss represents the mean and standard deviation of perplexity at $65,536$ steps. The final perplexity is reported at the end of pre-training ($524,288$ steps). SGLUE refers to SuperGLUE and WebQ refers to WebQuestions dataset. We report average, ROUGE-2, accuracy, and BLEU score for SuperGLUE, XSum, WebQuestions, and WMT EnDe, respectively, on the validation sets. \textbf{Note:} Results on WMT English to German are reported \textbf{without any pre-training}. The scores which outperform the vanilla Transformer are highlighted in \textbf{boldface}.}
    \label{tab:rel-attn-results}
\end{table*}



\subsection{Impact of hyperparameter tuning}
\label{sec:hparam-tuning}

It is a well-established fact in deep learning that hyperparameters (and even random seeds \citep{dodge2020finetuning}) may have a huge impact on model quality.
In our experiments, we intentionally kept hyperparameter fixed in order to measure whether a given modification improves performance regardless of hyperparameter settings.
Given that this may be an overly idealistic constraint, we present a case study of trying to improve one of the model variants by tuning its hyperparameters. We selected Universal Transformers (UT) \citep{dehghani2018universal} because it was claimed to achieve better results than the vanilla Transformer, and the UT has a relatively large number of hyperparameters that we can adjust. Using our standard hyperparameters, we obtain a loss of $2.40$ after training for $65{,}536$ steps. Bearing in mind that our vanilla Transformer obtains a loss of $2.182$ after the same amount of training, our goal was to at least achieve comparable performance using the UT. 

To this end, we swept over $25$ model configurations, varying the number of recurrent steps and the gating/transition functions in the UT. We also varied non-model-specific hyperparameters including the learning rate schedule and $d_{\rm{model}}$. Over these 25 sweeps, only $2$ managed to outperform the initial results. The only settings that worked were the result of reducing the number of recurrent steps (from $16$ to $2$) and slightly increasing the model size. In the end, we managed to achieve an improvement of $2.40 \rightarrow 2.265$ (or $6\%$ relative). While this is significant, many other hyperparameter settings failed to produce good results, and we were ultimately unable to match the performance of the vanilla Transformer. This exercise illustrates the challenge of tuning these models.

\begin{figure*}[t]
  \centering
  \subfigure[SuperGLUE]{\includegraphics[width=0.3\linewidth]{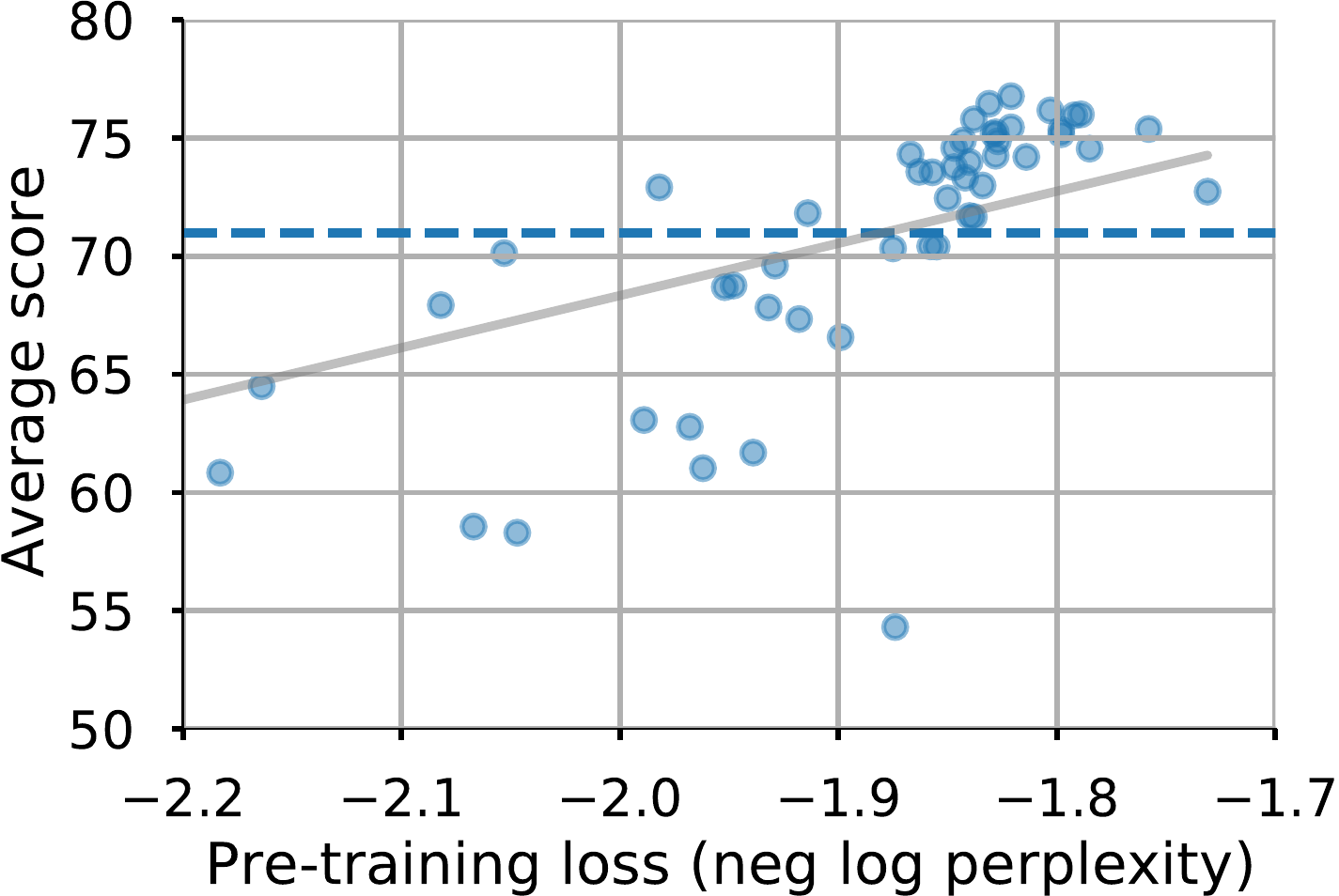}\label{fig:ppl-vs-avg}}\quad
  \subfigure[XSum]{\includegraphics[width=0.3\linewidth]{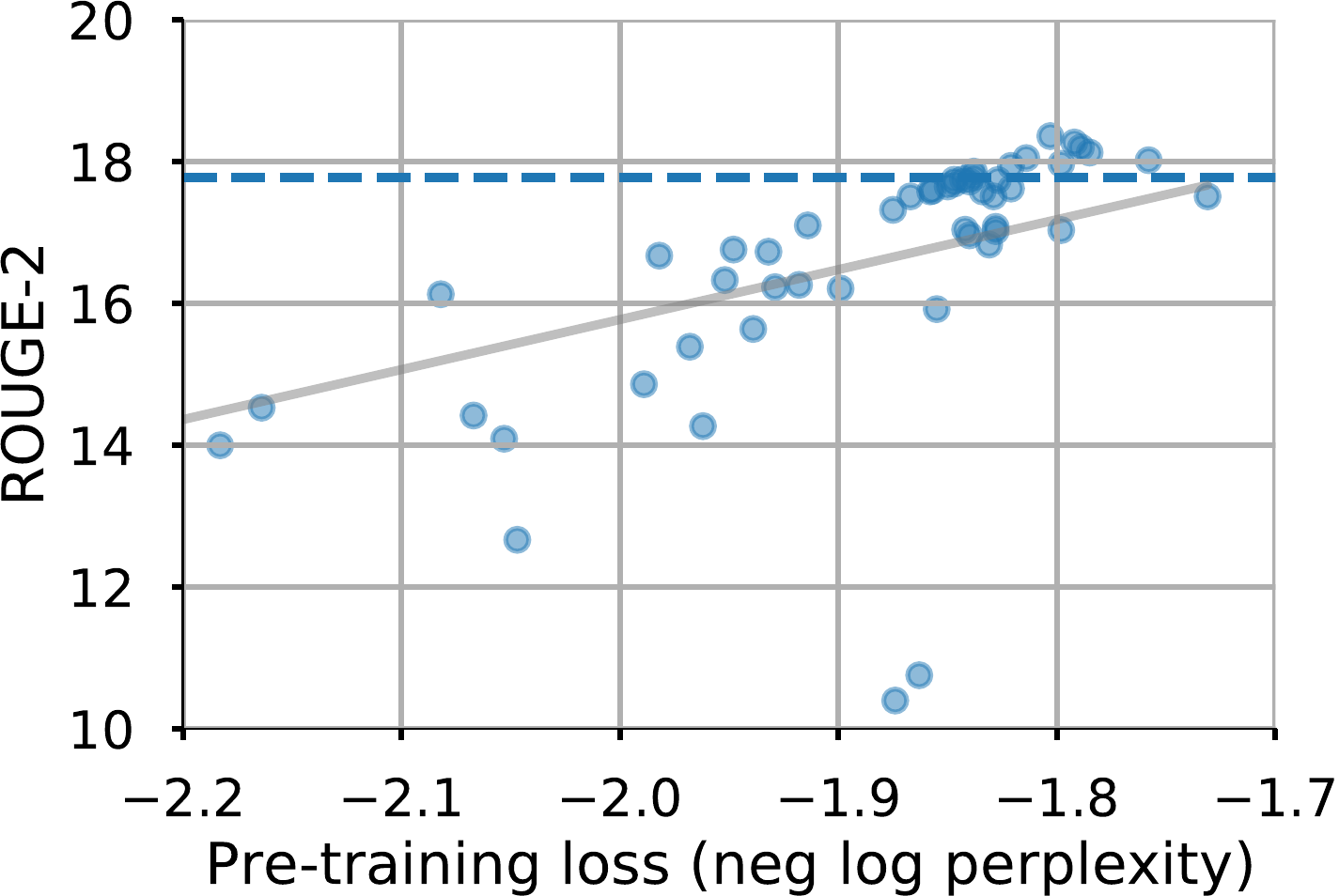}}\quad    
  \subfigure[WebQuestions]{\includegraphics[width=0.3\linewidth]{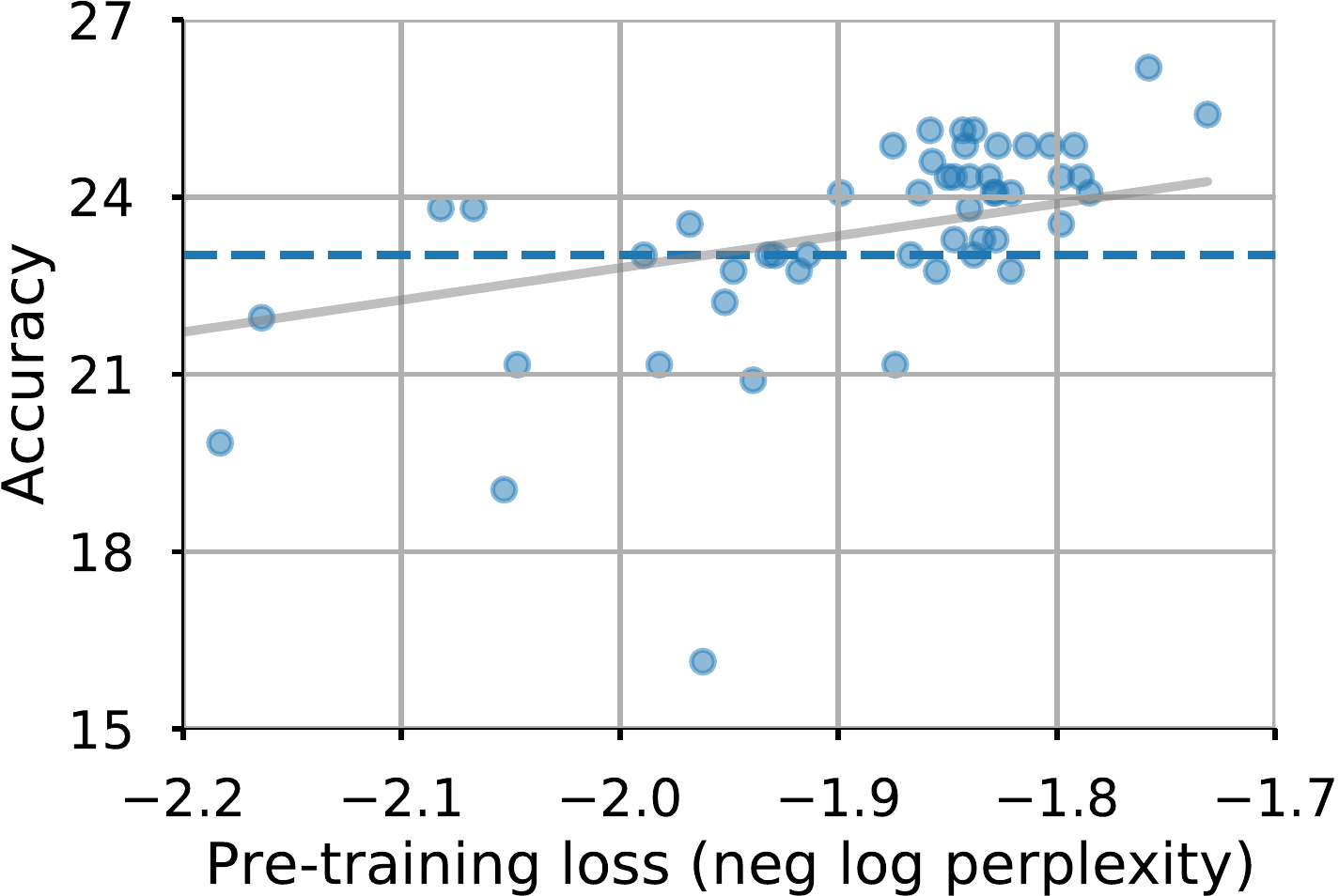}}
  \caption{Relationship between perplexity and fine-tuned task quality. The x-axis measures the pre-training perplexity and the y-axis measures the score for each task, with each point representing an architecture variant. The dashed line shows baseline performance and the gray line is the line of best fit.}
  \label{fig:ppl_correlation}
  \vspace{-10pt}
\end{figure*}

\vspace{-5pt}
\subsection{Correlation of perplexity and task performance}

In order to understand the relationship between pre-training performance and fine-tuned task quality, we investigate the correlation between perplexity and quality on each task. As shown in \cref{fig:ppl_correlation}, quality on all three tasks seem to be correlated with pre-training perplexity, though the correlation is surprisingly weak given past results suggesting a stronger relationship \cite{adiwardana2020towards}. Interestingly, the performance on SuperGLUE (Spearman's $\rho = 0.87$) and XSum (Spearman's $\rho = 0.80$) seems to be highly correlated with the pre-training perplexity, whereas the performance on WebQuestions (Spearman's $\rho = 0.69$) has a somewhat lower correlation. This may indicate that classification and generation tasks benefit more from improvements in perplexity than knowledge-intensive tasks like question answering. 



\section{Conjectures and Recommendations}
\label{sec:conjectures}

\vspace{-5pt}

As discussed above, we were surprised to find that so few of the architectural modifications produced improvements in the settings we considered.
This largely contrasts the experiments included in the original papers that proposed each modification.
We broadly grouped the modifications that actually did improve performance as either 1) being relatively simple (e.g.\ a change in activation function), 2) being developed in the same codebase where we ran experiments (e.g.\ the Synthesizer variants \cite{tay2020synthesizer}), or 3) incurring an increase in parameter count or FLOPs (e.g.\ the Switch Transformer \cite{fedus2021switch} or Universal Transformer \cite{dehghani2018universal}).
Other modifications that don't fit into one of these categories generally didn't improve performance.
There are various possible explanations as to why our results bore out the way they did:
\begin{enumerate}[topsep=0pt,itemsep=0ex,partopsep=1ex,parsep=1ex,wide, labelwidth=!,labelindent=0pt]
    \item \emph{The Mesh TensorFlow codebase and implementation are just so different than standard practice that most architectural modifications do not work.} We believe this is unlikely due to the fact that the Mesh TensorFlow Transformer implementation was created by one of the co-authors of the original Transformer paper and has been used to attain state-of-the-art results (e.g.\ \citet{raffel2019exploring,roberts2020much,khashabi2020unifiedqa,kale2020text,nogueira2020document,narang2020wt5,xue2020mt5,fedus2021switch}, etc.).
    \item \emph{The tasks we consider are non-standard or do not match the set of tasks used to vet the modifications in the first place}.
    The Transformer model is used for a variety of NLP problems including classification and generation tasks. We included transfer learning experiments on SuperGLUE, XSum, and WebQuestions and supervised training on WMT'14 EnDe, which covers the majority of use-cases. 
    \item \emph{Not tuning hyperparameters handicapped other methods}. While per-modification tuning might improve results (as verified in \cref{sec:hparam-tuning}), we argue that truly useful improvements to the Transformer should be reasonably hyperparameter-agnostic. Further, if hyperparameter sensitivity was the issue, it would be likely that a least a few of the compared methods ``got lucky'' with the hyperparameters, but very few modifications produced a boost.
    \item \emph{We implemented many of the modifications incorrectly}. To rule out this possibility, we corresponded with many of the creators of the modifications we considered, who confirmed the correctness in all cases.
    \item \emph{Modifications to the Transfomer architecture often do not transfer across implementations and applications}.
\end{enumerate}

Following the above rationale, we believe the final option is a plausible explanation for our results.
This possibility is supported by the fact that few of the modifications we consider in this paper have seen widespread adoption -- if they transferred easily across implementations and applications, they would likely have been more widely adopted.

Given this sober take, we conclude our paper with some suggestions as to how to ensure the robustness of improvements for future architectural modifications.
First, when proposing a new modification, try it out in multiple completely disparate codebases.
Given the proliferation of Transformer implementations (e.g.\ \citet{wolf2019huggingface,shazeer2018mesh,vaswani2018tensor2tensor}, etc.), this should be straightforward.
Second, apply it to a wide variety of downstream applications, including transfer learning, supervised learning, and language modeling -- and, possibly, include domains beyond NLP too (e.g., computer vision \cite{dosovitskiy2020image}).
Third, when evaluating performance in different implementations and on different tasks, keep hyperparameters fixed as much as possible, or at least attempt to measure the robustness of the modifications to changes in hyperparameters.
Finally, best-practice reporting of results should include mean and standard deviation across multiple trials, or at least avoid cherry-picking the best run \cite{dodge2020finetuning,henderson2018deep}.
With these guidelines in mind, we hope future work on architectural modifications to the Transformer will be more likely to see widespread adoption and improve the performance of this powerful architecture.

\bibliography{refs}
\bibliographystyle{acl_natbib}

\clearpage
\appendix
\section{Experiments with positional embeddings}
\label{sec:pos-embedding}

We also conducted a study of architectural variants using learned positional embeddings \citep{vaswani2017attention} in the baseline model instead of relative attention. Besides this change, the experimental setup remains the same (as described in \cref{sec:experiments}). The weighted Transformer architecture doesn't reliably converge using positional embeddings, so we do not report results using this architecture. 

In addition to the modifications described in \cref{sec:modifications}, we also experiment with variations in attention. Sinusoidal positional embeddings~\citep{vaswani2017attention} were proposed in the original Transformer to inject information of the order of the sequence into what was otherwise a set-operation transformation. Relative attention~\citep{shaw2018self} replaced the absolute position embeddings by those based on relative distance between tokens (clipped to a maximum distance hyperparameter $k$). 
The MeshTensorflow code base \cite{shazeer2018mesh} introduces two changes to relative attention.
In these changes, a bias is added to the self-attention logits (eq. \ref{eqn:encoder_self_attention}) before multiplication with values, where the bias may be optionally shared across self-attention layers.

The results from this study are shown in \cref{tab:pos-encoding-results}. Similar to relative attention, the only modifications that result in improvements are relatively minor modifications (e.g.\ activation function andnormalization), inefficient in terms of parameter count or FLOPs (e.g.\ the Switch Transformer) or were invented in the same codebase that we used (e.g.\ Synthesizer). Architectures with relative attention outperform those with positional embedding by a significant margin. Interestingly, certain architectures (Mixture of Softmaxes, tied decoder input and output embeddings) outperformed the vanilla Transformer with relative attention perform worse than the vanilla Transformer in this setup. Also, the absolute fine-tuned performance is worse for almost all the models compared with their relative attention counterparts. 

\begin{table*}[ht!]
    \centering
    \resizebox{\textwidth}{!}{%
    \begin{tabular}{p{6.0cm}cccccccc}
    \toprule
    \textbf{Model} & \textbf{Params} & \textbf{Ops} & \textbf{Step/s} & \textbf{Early loss} & \textbf{Final loss} & \textbf{SGLUE} & \textbf{XSum} & \textbf{WebQ} \\
    \midrule
    Vanilla Transformer & $223M$ & $11.1T$ & $3.90$ & $2.245 \pm 0.005$ & $1.865$ & $69.72$ & $16.94$ & $24.60$ \\ 
    \midrule
    GeLU &     $223M$ & $11.1T$ & $3.88$ & $2.220 \pm 0.005$ & $\textbf{1.863}$ & $\textbf{70.36}$ & $\textbf{17.10}$ & $23.28$  \\
    Swish &    $223M$ & $11.1T$ & $3.93$ & $2.234 \pm 0.005$ & $1.865$ & $\textbf{69.60}$ & $\textbf{17.07}$ & $24.34$ \\  
    ELU &      $223M$ & $11.1T$ & $3.86$ & $2.333 \pm 0.013$ & $1.942$ & $64.30$ & $16.21$ & $24.07$ \\    
    GLU &      $223M$ & $11.1T$ & $3.88$ & $2.212 \pm 0.005$ & $\textbf{1.834}$ & $\textbf{70.43}$ & $\textbf{17.42}$ & $24.34$ \\ 
    GeGLU &    $223M$ & $11.1T$ & $3.85$ & $2.172 \pm 0.010$ & $\textbf{1.807}$ & $\textbf{72.36}$ & $\textbf{17.69}$ & $\textbf{24.87}$ \\ 
    ReGLU &    $223M$ & $11.1T$ & $3.87$ & $2.190 \pm 0.008$ & $\textbf{1.832}$ & $\textbf{70.63}$ & $\textbf{17.38}$ & $21.96$\\ 
    SeLU &     $223M$ & $11.1T$ & $3.84$ & $2.372 \pm 0.016$ & $1.967$ & $64.68$ & $16.00$ & $23.28$ \\  
    SwiGLU &   $223M$ & $11.1T$ & $3.82$ & $2.168 \pm 0.006$ & $\textbf{1.806}$ &  $\textbf{70.90}$ & $\textbf{17.51}$ & $\textbf{25.13}$ \\ 
    LiGLU &    $223M$ & $11.1T$ & $3.88$ & $2.180 \pm 0.002$ & $\textbf{1.816}$ & $\textbf{71.23}$ & $\textbf{17.55}$ & $\textbf{24.60}$\\ 
    Sigmoid &  $223M$ & $11.1T$ & $3.94$ & $2.947 \pm 1.152$ & $1.908$ & $69.36$ & $16.64$ & $23.02$\\  
    Softplus & $223M$ & $11.1T$ & $3.77$ & $2.324 \pm 0.032$ & $1.885$ & $68.99$ & $16.92$ & $21.96$ \\  
    \midrule
    RMS Norm           & $223M$ & $11.1T$ & $3.99$ & $2.209 \pm 0.008$ & $\textbf{1.856}$ & $69.11$ & $16.90$ & $23.55$ \\
    Rezero             & $223M$ & $11.1T$ & $4.14$ & $3.180 \pm 0.719$ & $2.506$ & $54.01$ & $6.44$ & $20.90$ \\
    Rezero + LayerNorm & $223M$ & $11.1T$ & $3.78$ & $2.229 \pm 0.006$ & $1.902$ & $64.75$ & $16.40$ & $23.02$ \\
    Rezero + RMS Norm  & $223M$ & $11.1T$ & $3.90$ & $2.306 \pm 0.016$ & $1.948$ & $59.86$ & $15.66$ & $23.02$ \\
    Fixup              & $223M$ & $11.1T$ & $3.32$   & $2.473 \pm 0.014$ & $2.236$ & $57.98$ & $12.51$ & $23.28$ \\ 
    \midrule
    24 layers, $d_{\rm{ff}} = 1536, H = 6$ & $224M$ & $11.1T$ & $3.12$ & $2.260 \pm 0.014$ & $1.874$ & $\textbf{70.59}$ & $\textbf{17.11}$ & $23.02$ \\
    18 layers, $d_{\rm{ff}} = 2048, H = 8$  & $223M$ & $11.1T$ & $3.27$ & $2.268 \pm 0.037$ & $1.878$ & $\textbf{70.40}$ & $16.87$ & $23.02$ \\
    8 layers, $d_{\rm{ff}} = 4608, H = 18$  & $223M$ & $11.1T$ & $3.61$ & $2.243 \pm 0.003$ & $1.871$ & $68.67$ & $\textbf{17.03}$ & $23.55$ \\
    6 layers, $d_{\rm{ff}} = 6144, H = 24$  & $223M$ & $11.1T$ & $3.59$ & $2.250 \pm 0.004$ & $1.882$ & $68.08$ & $16.93$ & $23.81$ \\
    \midrule
    Block sharing                     & $65M$ & $11.1T$ & $4.03$ & $2.777 \pm 0.019$ & $2.237$ & $63.06$ & $13.89$ & $21.96$ \\
    \hspace*{0.1cm} + Factorized embeddings           & $45M$ & $9.4T$ & $4.35$ & $2.670 \pm 0.178$ & $2.205$ & $57.17$ & $12.13$ & $20.11$ \\
    \hspace*{0.1cm} + Factorized \& Shared embeddings               & $20M$ & $9.1T$ & $4.49$ & $2.874 \pm 0.059$ & $2.362$ & $57.46$ & $11.78$ & $19.58$ \\
    Encoder only block sharing        & $170M$ & $11.1T$ & $3.80$ & $2.399 \pm 0.008$ & $2.016$ & $64.08$ & $14.74$ & $21.69$ \\
    Decoder only block sharing        & $144M$ & $11.1T$ & $3.92$ & $2.542 \pm 0.067$ & $2.048$ & $\textbf{69.95}$ & $16.01$ & $21.96$ \\ 
    \midrule
    Factorized Embedding                & $227M$ & $9.4T$  & $3.97$ & $2.273 \pm 0.019$ & $1.886$ & $68.91$ & $16.41$ & $21.43$ \\
    Factorized \& shared embeddings     & $202M$ & $9.1T$  & $4.08$ & $2.387 \pm 0.006$ & $2.018$ & $\textbf{69.93}$ & $16.07$ & $21.96$ \\
    Tied encoder/decoder input embeddings   & $248M$ & $11.1T$ & $3.86$ & $2.254 \pm 0.008$ & $1.872$ & $68.34$ & $16.60$ & $22.75$ \\
    Tied decoder input and output embeddings   & $248M$ & $11.1T$ & $3.86$ & $2.262 \pm 0.006$ & $1.871$ & $69.48$ & $16.85$ & $23.28$ \\
    Untied embeddings            & $273M$ & $11.1T$ & $3.83$ & $2.265 \pm 0.013$ & $1.872$ & $67.99$ & $16.66$ & $23.02$ \\    
    Adaptive input embeddings                      & $204M$ & $9.2T$  & $4.15$ & $2.321 \pm 0.006$ & $1.934$  & $69.20$ & $16.69$ & $21.96$ \\  
    \midrule
    Adaptive softmax                    & $204M$ & $9.2T$  & $4.21$ & $2.425 \pm 0.005$ & $2.009$  & $67.71$ & $15.74$ & $20.11$ \\
    Adaptive softmax without projection & $223M$ & $10.8T$ & $3.97$ & $2.357 \pm 0.009$ & $1.937$ & $68.68$ & $16.45$ & $22.75$ \\
    Mixture of softmaxes                & $232M$ & $16.3T$ & $2.50$ & $3.112 \pm 1.169$ & $\textbf{1.843}$ & $\textbf{70.70}$ & $16.78$ & $22.75$ \\
    \midrule    
    Relative attention with bias        & $223M$ & $11.3T$ & $3.49$ & $2.197 \pm 0.005$ & $\textbf{1.832}$ & $\textbf{74.06}$ & $\textbf{17.63}$ & $\textbf{24.87}$ \\
    Relative attention with shared bias & $223M$ & $11.3T$ & $3.57$ & $2.194 \pm 0.006$ & $\textbf{1.840}$ & $\textbf{74.14}$ & $\textbf{17.62}$ & $24.34$ \\
    Relative position representation    & $223M$ & $11.1T$ & $3.10$ & $2.189 \pm 0.008$ & $\textbf{1.838}$ & $\textbf{74.26}$ & $\textbf{17.67}$ & $24.07$ \\
    Sinusoidal positional encoding      & $223M$ & $11.1T$ & $3.91$ & $2.278 \pm 0.032$ & $1.906$ & $\textbf{69.76}$ & $16.25$ & $22.75$  \\
    Transparent attention               & $223M$ & $11.1T$ & $3.61$ & $2.244 \pm 0.013$ & $1.949$ & $53.77$ & $6.39$ & $15.08$ \\
    Dynamic convolution                 & $257M$ & $11.8T$ & $2.65$ & $2.405 \pm 0.007$ & $2.038$ & $55.16$ & $10.25$ & $4.50$ \\  
    Lightweight convolution             & $224M$ & $10.4T$ & $4.05$ & $2.356 \pm 0.006$ & $1.990$ & $61.32$ & $14.08$ & $24.08$ \\    
    Evolved Transformer                 & $217M$ &  $9.7T$ & $3.11$ & $2.233 \pm 0.004$ & $1.890$ & $67.88$ & $16.40$ & $24.08$ \\ 
    Synthesizer (dense)                 & $224M$ & $11.4T$ & $3.61$ & $2.339 \pm 0.019$ & $1.965$ & $61.02$ & $14.48$ & $18.25$ \\ 
    Synthesizer (dense plus)            & $243M$ & $12.6T$ & $3.34$ & $2.200 \pm 0.008$ & $\textbf{1.832}$ & $\textbf{74.16}$ & $\textbf{16.96}$ & $\textbf{24.87}$ \\ 
    Synthesizer (dense plus alpha)      & $243M$ & $12.6T$ & $3.11$ & $2.204 \pm 0.005$ & $\textbf{1.846}$ & $\textbf{75.18}$ & $16.94$ & $\textbf{24.60}$ \\  
    Synthesizer (factorized)            & $207M$ & $10.1T$ & $4.10$ & $2.629 \pm 0.573$ & $1.964$ & $61.76$ & $15.44$ & $22.49$ \\ 
    Synthesizer (random)                & $254M$ & $10.1T$ & $4.26$ & $2.458 \pm 0.167$ & $1.972$ & $64.61$ & $15.39$ & $23.02$ \\ 
    Synthesizer (random plus)           & $292M$ & $12.0T$ & $3.79$ & $2.202 \pm 0.010$ & $\textbf{1.849}$ & $\textbf{76.84}$ & $\textbf{17.04}$ & $23.02$ \\  
    Synthesizer (random plus alpha)     & $292M$ & $12.0T$ & $3.55$ & $2.212 \pm 0.013$ & $\textbf{1.856}$ & $\textbf{75.02}$ & $\textbf{17.08}$ & $\textbf{24.87}$ \\
    Universal Transformer               & $84M$  & $40.0T$ & $0.88$ & $2.443 \pm 0.022$ & $2.111$ & $60.54$ & $12.02$ & $17.73$ \\ 
    Mixture of experts                  & $648M$ & $11.7T$ & $3.20$ & $2.194 \pm 0.008$ & $\textbf{1.846}$ & $68.82$ & $\textbf{17.12}$ & $\textbf{24.87}$ \\ 
    Switch Transformer                  & $1100M$ & $11.8T$ & $3.41$ & $2.175 \pm 0.005$ & $\textbf{1.775}$ & $\textbf{72.21}$ & $\textbf{17.78}$ & $\textbf{24.87}$ \\  
    Funnel Transformer                  & $223M$ & $1.9T$ & $4.83$  & $2.291 \pm 0.008$ & $1.925$ & $67.11$ & $16.33$ & $21.64$ \\ 
    Product key memory                  & $421M$ & $386.6T$ & $0.25$ & $2.212 \pm 0.007$ & $\textbf{1.821}$ & $69.62$ & $16.58$ & $24.08$ \\  
    \bottomrule
    \end{tabular}}
    \caption{Pre-training and fine-tuning results for all architecture variants with \emph{learned positional embeddings}. The early loss represents the mean and standard deviation of perplexity at $65,536$ steps. The final perplexity is reported at the end of pre-training ($524,288$ steps). SGLUE refers to SuperGLUE and WebQ refers to WebQuestions dataset. We report average, ROUGE-2, and accuracy for SuperGLUE, XSum, and WebQuestions, respectively, on the validation sets. The scores which outperform the vanilla Transformer are highlighted in \textbf{boldface}.}
    \label{tab:pos-encoding-results}
\end{table*}

\section{Implementation details for modifications}
\label{sec:implementation-details}

For factorized embedding, we use an inner dimension of $128$ for models with and without block sharing of parameters.

In adaptive input embedding experiments, we use three clusters of size $2500$, $6000$, and $23,628$. For experiments with adaptive softmax, we split the third cluster into two clusters of $23,500$ and $128$. Since we used a larger vocabulary (see section~\ref{sec:experiments}) for the supervised training on the WMT'14, we use the same number of clusters with the same relative cluster sizes.

We experimented with $10$ and $15$ softmaxes for the mixture of softmax models. In the paper, we only report results for the model with $15$ softmaxes since it performs better. 

For Lightweight and Dynamic convolutions, we use one-dimensional kernel with width 9. The depth of the kernel is determined depending on whether it is depthwise-convolution or vanilla convolution in which case its depth is $d_{\rm{model}}$. For Universal Transformer, we use number of recurrent steps of $24$ and halting threshold of $0.5$. We use $32$ experts in the Mixture of Experts experiments. 

In PKM experiments, we use $knn = 32$, $128$ keys and $512$ memory slots. In our experiments, we introduce a product key memory network before the last layer in the decoder.

In the Funnel Transformer experiments, we use mean pooling with $3$ blocks in the encoder. The input sequence is pooled after every $4$ layers in the funnel Transformer. In the weighted Transformer, we freeze the weights of the branched attention module for the last $20,000$ steps of pre-training.

\section{Reproducing the original Transformer experiments}
\label{sec:aiayn_reprod}

\citet{vaswani2017attention} reported the BLEU score of 25.8 (Table 3 of their paper) when evaluated on the dev set without checkpoint averaging. We ran a replication experiment with the same Transformer-Base architecture and achieved 25.52. With this, we believe that our Transformer codebase closely replicates the original one. Additionally, the baseline transformer model in our paper is comparable to the Transformer-Big model from~\citet{vaswani2017attention}. The Transformer-Big model achieves a BLEU score of 26.4 (Table 3 of their paper) on the validation set of the WMT EnDe translation task. Our baseline model achieves a BLEU score of 26.62 on the same validation set which is marginally better than the results reported in the original paper.

\section{Transformer Background}
\label{sec:background}

In this section, we give a brief description of the original Transformer architecture. We primarily include this description so that we can refer back to specific components as we introduce different modifications.
For a more in-depth description of the Transformer architecture, refer to the original paper \citep{vaswani2017attention} or follow-up tutorials\footnote{\url{http://nlp.seas.harvard.edu/2018/04/03/attention.html}}\textsuperscript{,}\footnote{\url{http://jalammar.github.io/illustrated-transformer/}}.

In this work, we solely experiment with ``encoder-decoder'' Transformers, which ingest an input sequence of tokens and produce an output sequence conditioned on the input.
We denote the tokens of the input sequence as $x[1], x[2], \ldots, x[T]$ and the target sequence as $y[1], y[2], \ldots, y[U]$.
The encoder first embeds each entry in the input sequence using the embedding matrix $E \in \mathbb{R}^{d_{\mathrm{vocab}} \times d_{\mathrm{model}}}$ and adds a position encoding $p$ as follows:
\begin{equation*}
h_{e, 0}[t] = E[x[t]] + p[t]
\end{equation*}
where $p[t] \in \mathbb{R}^{d_{\mathrm{model}}}$ is a ``position embedding''.
In the original Transformer, this position embedding is computed as
\begin{equation}
    p[t, i] =
    \begin{cases}
    \sin\left(\frac{t}{10000^{2i/d_{\mathrm{model}}}}\right) & i\;\mathrm{even} \\[10pt]
    \cos\left(\frac{t}{10000^{2i/d_{\mathrm{model}}}}\right) & i\;\mathrm{odd}
    \label{eq:pos-sinusoidal}
\end{cases}
\end{equation}
In general, we will use $h_{e, l}$ and $h_{d, l}$ to denote the output of the $l$th layer block of the encoder and decoder, respectively.
For simplicity, we refer to the embeddings as if they are the output of a ``zeroth'' layer block.

Each layer block in the encoder comprises a multi-headed self-attention mechanism  \cite{cheng2016long} followed by a position-wise dense/nonlinearity/dense feedforward network.
Both of these ``subblocks'' include a residual connection \cite{he2016deep} and layer normalization \cite{ba2016layer}.
Layer normalization is defined as an operation over a sequence $h[1], \ldots, h[T]$ as
\begin{align}
    \mu[t] &= \frac{1}{d_{\mathrm{model}}} \sum_{i = 1}^{d_\mathrm{model}} h[t, i] \\
    \sigma[t] &= \sqrt{ \frac{1}{d_{\mathrm{model}}} \sum_{i = 1}^{d_\mathrm{model}} (h[t, i] - \mu[t])^2} \\
    \layernorm(h)[t] &= \frac{\gamma}{\sigma[t]} \odot (h[t, i] - \mu[t]) + \beta
\end{align}
where $\odot$ indicates elementwise multiplication and $\gamma, \beta \in \mathbb{R}^{d_{\mathrm{model}}}$ are learned parameters that are unique to each instance of layer normalization.

Head $h$ in the multi-headed self-attention of layer $l$ produces, at timestep $t$,
\begin{align}
    q_{e, l, h}[t] &= h_{e, l - 1}[t]Q_{e, l, h} \\
    k_{e, l, h}[t] &= h_{e, l - 1}[t]K_{e, l, h} \\
    v_{e, l, h}[t] &= h_{e, l - 1}[t]V_{e, l, h} \\
    a_{e, l, h} &= \softmax\mathopen{}\left(\frac{q_{e, l, h}[t] k_{e, l, h}[t]^\top} {\sqrt{d_k}}\right)v_{e, l, h}[t]
    \label{eqn:encoder_self_attention}
\end{align}
where $Q_{e, l, h} \in \mathbb{R}^{d_{\mathrm{model}} \times d_k}$, $K_{e, l, h} \in \mathbb{R}^{d_{\mathrm{model}} \times d_k}$, and $V_{e, l, h} \in \mathbb{R}^{d_{\mathrm{model}} \times d_v}$ are the ``query'', ``key'', and ``value'' projection matrices, respectively.
The self-attention outputs $a_{e, l, h}$ for all $H$ heads are then concatenated and projected against the matrix $O_{e, l} \in \mathbb{R}^{Hd_v \times d_\mathrm{model}}$ along with a residual connection and layer normalization as follows:
\begin{equation}
    s_{e, l}[t] = \layernorm\mathopen{}\left(\begin{bmatrix}a_{e, l, 1}[t] \\ \vdots \\ a_{e, l, H}[t]\end{bmatrix}O_{e, l} + h_{e, l - 1}[t]\right) \label{eqn:residual}
\end{equation}

The output of the multi-headed self-attention mechanism is then passed through a feedforward network that operates on each sequence element independently.
Specifically, the feedforward network consists of a projection, a ReLU nonlinearity, and another projection as follows:
\begin{equation}
f_{e, l}[t] = \max(0, s_{e, l}[t] W_{e, l, 1} + b_{e, l, 1})W_{e, l, 2} + b_{e, l, 2} \label{eqn:ff}
\end{equation}
where $W_{e, l, 1} \in \mathbb{R}^{d_{\mathrm{model}} \times d_{\mathrm{ff}}}, b_{e, l, 1} \in \mathbb{R}^{d_{\mathrm{ff}}}, W_{e, l, 1} \in \mathbb{R}^{d_{\mathrm{ff}} \times d_{\mathrm{model}}}$ and  $b_{e, l, 1} \in \mathbb{R}^{d_{\mathrm{model}}}$.
The output of the feedforward network is then combined with the subblock's input via a residual connection and layer normalization:
\begin{equation}
    h_{e, l} = \layernorm(s_{e, l} + f_{e, l})
    \label{eqn:ln}
\end{equation}

Overall, the decoder is structured similarly to the encoder, with the following changes:
First, the self-attention mechanisms are ``causal'' which prevents the decoder from looking at future items from the target sequence when it is fed in during training.
This is achieved by constructing an ``attention mask'' $M \in \mathbb{R}^{U \times U}$ that zeros out attention entries that are nonpermissable; specifically replacing the operation in \cref{eqn:encoder_self_attention} with
\begin{align}
    M[i, j] &= \begin{cases}
    0, & i \le j\\
    -\infty, & i > j
    \end{cases}\\
    a_{d, l, h} &= \softmax\mathopen{}\left(\frac{q_{d, l, h}[t] k_{d, l, h}[t]^\top} {\sqrt{d_k}} + M\right)v_{d, l, h}[t]
    \label{eqn:decoder_self_attention}
\end{align}
where the $d$ subscript denotes activations and parameters for the decoder.
Second, the layer blocks in the decoder contain an encoder-decoder attention mechanism after the self-attention mechanism and before the feedforward network.
Specifically, encoder-decoder attention computes
\begin{align}
    q^\prime_{d, l, h}[t] &= s_{d, l}[t]Q^\prime_{d, l, h} \\
    k^\prime_{d, l, h}[t] &= h_{e, L}[t]K^\prime_{d, l, h} \\
    v^\prime_{d, l, h}[t] &= h_{e, L}[t]V^\prime_{d, l, h} \\
    a^\prime_{d, l, h} &= \softmax\mathopen{}\left(\frac{q^\prime_{d, l, h}[t] k^\prime_{d, l, h}[t]^\top} {\sqrt{d_k}}\right)v^\prime_{d, l, h}[t] \label{eqn:encoder_decoder_attention}
\end{align}
The activations from each head $a^\prime_{d, l, h}$ are then fed into the residual/layer norm block (\cref{eqn:residual}) and the feedforward network (\cref{eqn:ff}) as usual.

At the output of the final layer of the decoder, each entry in the sequence of activations $h_{d, L}$ is projected via an output logit matrix $G \in \mathbb{R}^{d_{\mathrm{model}} \times d_{\mathrm{vocab}}}$.

\end{document}